\documentclass[letterpaper]{article}
\usepackage[utf8]{inputenc}

\usepackage{graphicx}
\usepackage{amsfonts}
\usepackage{amsmath}
\usepackage{graphicx}
\usepackage{color}
\usepackage{authblk}

\title{Analyzing the Conjunction Fallacy as a Fact}
\author{Tomas Veloz$^{1,2,3}$\\tomas.veloz@vub.be \and Olha Sobetska$^{4}$\\
sobetska.olha@gmail.com}
\affil[1]{Centre Leo Apostel, Vrije Universiteit Brussel, Belgium}
\affil[2]{Mathematics Department, Metropoitan Technological University, Santiago, Chile}
\affil[3]{Interdisciplinary Foundation for the Development of Science, Technology and Arts}
\affil[4]{Faculty of Social Sciences and Philosophy, Institute of Sociology, University of Leipzig, Beethoven Street 15, 04107, Leipzig, Germany}
\begin{document}
\vspace{-4cm}
\maketitle

\begin{abstract}
Since the seminal paper by Tversky and Kahneman, the `conjunction fallacy' has been the subject of multiple debates and become a fundamental challenge for cognitive theories in decision-making. In this article, we take a rather uncommon perspective on this phenomenon. Instead of trying to explain the nature or causes of the conjunction fallacy (intensional definition), we analyze its range of factual possibilities (extensional definition). We show that the majority of research on the conjunction fallacy, according to our sample of experiments reviewed which covers literature between 1983 and 2016, has focused on a narrow part of the a priori factual possibilities, implying that explanations of the conjunction fallacy are fundamentally biased by the short scope of possibilities explored. The latter is a rather curious aspect of the research evolution in the conjunction fallacy considering that the very nature of it is motivated by extensional considerations. 


\end{abstract}
\medskip
{\bf Keywords: conjunction fallacy; possible experiences; factuality}

\maketitle

\section{Introduction\label{intro}}


The conjunction fallacy (CF) is one of the most important challenges in rational approaches to cognition \cite{hd2001,ggk2002}. It has been extensively discussed in the cognitive science community \cite{tk1983,mb1984,g1996,tbo2004,mo2009}. Tversky and Kahneman introduced this fallacy in their influential 'Linda story' experiment \cite{tk1983}, where people tend to judge the conjunction of two events $A$ and $B$ as more likely than $A$ or $B$ separately. For example, people judge ''Linda is a feminist and a bank teller'' as more likely than ''Linda is a bank teller''. The CF poses a challenge to classical probability modelling, and its presence has been confirmed in several cognitive experiments involving Linda-like stories \cite{mb1984,tbo2004,gre1991,fp1996,f2002,wm2008,c2009,Lu2015}. 

Tversky and Kahneman's developed a research program to explain this and other fallacies based on `individual biases and heuristics'. This work became extremely influential in decision-making theories and encouraged the development of several alternative explanations based on normative and descriptive approaches~\cite{bpft2011,Lu2015}. However, there is no agreement on the ultimate cognitive process that produces these deviations~\cite{so2008}.

\begin{figure}[htbp]
  \centering
  \includegraphics[width=0.5\textwidth]{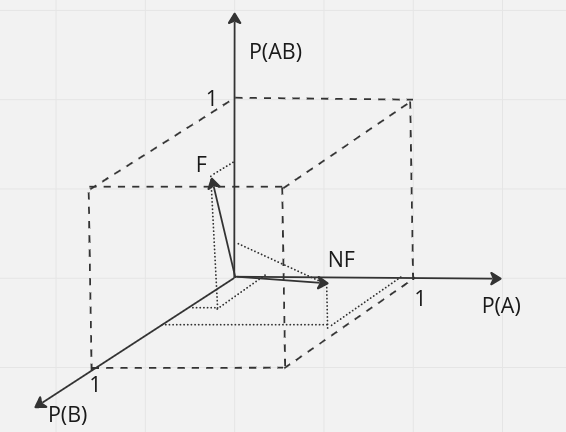}
  \caption{Examples of two data points, one not committing the fallacy ($NF$) and another committing it ($F$). Note that for $NF$, the projections of the vector onto the $P(A)$ and $P(B)$ axes are larger than the projection onto the axis $P(AB)$, while the opposite is true for $F$}
  \label{CFNCF}
\end{figure}

It should be noted that the CF requires evidence of no more than three numbers. Namely, the likelihood estimates of events '$A$', '$B$', and '$A \text{ and }  B$'. These likelihood estimates can in principle be any value between zero and one, where zero means completely unlikely and one means completely likely.
The CF is represented as a point in $[0,1]^3$, i.e. a three-dimensional vector $(P(A),P(B),P(AB))$ with values between $0$ and $1$ where the rational approach, inherited from Kolmogorovian probability theory or equivalently Fuzzy logic, expects that the third value of such a vector must be smaller or equal than the other two. Namely, if a likelihood estimate of events '$A$', '$B$', and '$A \text{ and } \ B$' produces a point is such that
\begin{equation}
P(AB)\leq \min(P(A),P(B)),
\label{CF0}
\end{equation}

then we say that the likelihood estimation commits the CF. In figure~\ref{CFNCF} we show the cube formed by $[0,1]^3$ and two vectors representing a case where the fallacy is not committed ($NF$), and another vector where the fallacy is committed ($F$).

Research on the CF has steadily developed for exactly four decades, and the seminal article (SA) of Tversky and Kahneman accumulates today more than 6000 citations. Regarding models explaining the CF, Tversky and Kahneman's model, as well as several other approaches, have produced formulas that resemble in some way our rational inequality eq.~\ref{CF0}, but by introducing other 'psychological' parameters, ensure that the third value can be larger than one or two of the other values~\cite{TCR2013,bketal2015}. 

Concerning experimental work on the CF, a large number of articles do not provide experimental evidence at the probability judgement level, but instead, report the percentage of people 'committing the fallacy'~\cite{fiedler1988dependence}. The latter prevents scholars in the community researching CF from comparing or replicating experimental data. Additionally, in several experiments other measures that are related to probability judgements such as likelihood ranking and preference are given~\cite{bar1984representativeness,agnoli1989suppressing}, while others report two of the three likelihood estimates, preventing proper comparison across experiments and parameters of alternative explanations~\cite{shafffi1990typicality}.

Due to the latter problems, we will discuss in this article the extensional aspect of the CF. Namely, we focus on the different possibilities that can arise from any given point in $[0,1]^3$ representing likelihood estimates of $A$, $B$, and $A \text{ and }  B$, and what can be inferred about the fallacy in the different regions. To do so, we reviewed nearly 4000 articles citing the CF up to 2016. Strikingly, We found that the literature has focused on estimates that cover a narrow part of 
$[0,1]^3$. To conclude, we will discuss why the latter has happened and how to advance this research program further.

\section{The conjunction fallacy in a nutshell}
\label{data}

In the field of cognitive psychology, Tversky and Kahneman discovered the CF through an experiment known as the "Linda story" \cite{tk1983}. Participants received a questionnaire that featured a story about a woman named Linda who was a single, outspoken, and bright philosophy major. Afterwards, participants were asked to estimate the likelihood of several options, where two of them were: (i) Linda is a bank teller or (ii) Linda is a bank teller and is active in the feminist movement. The experiment found that 85\% of the respondents chose option (ii) over option (i) \cite{tk1983}. This result contradicts the classical Kolmogorovian probability, which predicts that $P(A \ {\rm and} \ B) \le P(B)$.

In the SA by Tversky and Kahneman, they performed a number of experiments testing the CF in other Linda-like situations including outcomes of dice rolling, features or decisions of people in specific situations, potential outcomes of a diagnosis, competition results in a tournament, etc. In their experiments, different types of estimates were considered, including likelihood, probability, ranking, etc. Moreover, different considerations regarding the level of general and specific knowledge of participants were also explored. The important and striking result is that, in all tested situations, CF was observed in a significant part of the studied group. As an explanatory mechanism, they proposed the representativeness heuristic. That is, the event "A and B" is more representative than B alone \cite{bpft2011}. In the words of Tversky and Kahneman, {\it The representativeness heuristic favours outcomes that make good stories or good hypotheses}'~\cite{tk1983}. The representativeness heuristic is theoretically developed in the company of various other heuristics based on concepts such as availability, anchoring, causal coherence, etc., which all together form a complex of notions proposing a theory for non-rational decision-making, which proved quite useful and led Kahneman to obtain the Nobel prize in economics in 2002~\cite{shefrin2003contributions}. 

While a myriad of studies appeared after SA confirming the CF in multiple other situations~ \cite{mb1984,tbo2004,gre1991,f2002,wm2008,c2009}, a few approaches were proposed to explain the fallacy in a more formal-theory-friendly style. The "misunderstanding hypothesis" is the most important of such explanations. It suggests that people fail to understand the meaning of sentences and get confused due to linguistic interpretations. For example, the misunderstanding hypothesis has proposed that when we read options 'Linda is a bank teller' and 'Linda is a feminist and a bank teller' we might tend to think that the former option implies 'Linda is not a feminist'~\cite{fiedler1988dependence}. Various mathematical models have tried to explain either the misunderstanding hypothesis or some semantic processing-inspired version of how people deal with likelihood estimates of individual events to create a likelihood estimate of a compound event. However, experimental challenges have also been raised against these models~\cite{tbo2004,wm2008}, leaving the CF as one of the cornerstones of non-rational decision-making.

Beyond the purely decision-making aspect, the CF has been explored from multiple other perspectives, such as neuropsychology, by measuring neural mechanisms that might underlie the decision-making process leading to a fallacy~\cite{gardner2019optimality}, by developing cross-cultural studies to understand whether or not the CF is somehow culture dependent~\cite{lee2018cross}, by detecting analogous fallacies for the case of disjunction and other logical connectives~\cite{mb1984,bb2012}, among others (for a review, see~\cite{mo2009}).

\section{Is the conjunction fallacy a fact?}
We recall that the CF is represented as a point in $[0,1]^3$, i.e., a three-dimensional vector $(P(A),P(B),P(AB))$ with values between $0$ and $1$ (see fig.\ref{CFNCF}). Therefore, any point in $[0,1]^3$ should {\it a priori} be a possible fact observed in an experimental setting.  

Probability theory teaches us that if we represent $A$ and $B$ as sets, and $P(A)$ and $P(B)$ reflect the measure of these sets, then not every point in $[0,1]^3$ can be a valid fact because the nature of set operations prevents $P(AB)$ from being larger than both $P(A)$ and $P(B)$. The above-mentioned assumptions are quite reasonable because every macroscopic physical event we observe obeys such rules. Indeed, let us consider a typical example of probability theory to illustrate this idea. Consider an urn with balls that can be either $A=$'red' or $A'=$'not red', and can be made either $B=$'wooden' or $B'=$'not wooden'. It is clear that the probability of extracting a ball that is 'red and wooden', i.e $P(AB)$ is smaller than extracting a ball that is 'red' ($P(A)$) or 'wooden' ($P(B)$). Additionally, other probabilistic inequalities can be assumed to hold without any risk, such as $1+P(AB)-P(A)-P(B)\geq 0$, and $P(A)+P(B)-P(A\cup B)\leq 1$, where $P(A\cup B)$ is the probability that the ball is 'red or wooden.'

Inspired by this reasoning, George Boole constrained the possible relative frequencies of experimental settings and formulated, for the first time in history, a set of probabilistic inequalities derived from set-theoretical considerations, and called them 'conditions of possible experience'~\cite{boole1854investigation}. The conditions of possible experience underlie all reasoning about our macroscopic physical reality and are fundamental in areas such as statistical mechanics and electrodynamics~\cite{cochrane2006kant}. However, it is today well-known that the conditions of possible experience fail for microscopic physical systems, specifically for those where quantum theory operates. The double-slit experiment is the canonical piece of evidence illustrating that quantum systems are 'not urns with balls' but something much stranger~\cite{wootters1979complementarity}. In the experiment, a beam of particles, such as electrons, is fired at a barrier with two slits. On the other side of the barrier, a detector is placed to detect where the particles land. The particles typically create an interference pattern on the detector screen, as if the particles were behaving as waves. The latter defies our intuition of particles behaving as 'balls passing through the slit' and seems to suggest that each particle 'passes through both slits simultaneously' as if particles were waves. This phenomenon is known as the wave-particle duality of matter, and it remains a cornerstone of our understanding of the behaviour of matter and energy at the quantum level.

A prominent philosopher of quantum physics called Itamar Pitowski studied why the conditions of possible experience are not respected by quantum systems and came up with a list of reasons that could cause violations to them~\cite{pitowsky1994george}:   

\begin{enumerate}
\item Failure of randomness: Obtaining a bad approximation of probabilities due to not taking a large enough sample of measurements.
\item Measurement biases: The method of experimentation produces disturbances in the object we measure with.
\item No distribution: The phenomena under observation does not possess a well-defined distribution of properties to be measured, or not even well-defined properties to think of a distribution in the first place. 
\item Mathematical oddities: Sets representing events might not be measurable, implying that probabilities will not be uniquely defined. 
\end{enumerate}

Pitowski explains that we do not know in principle what of these reasons could be in operation as a source of violation of the conditions of possible experience in quantum physics, though the first reason can be discarded due to the tremendous advances in experimental microphysics~\cite{schirber2022nobel}, and that reason four focuses on the mathematical structure of events to explain the violation of probabilities rather than on the very factual nature of the events and their observation. Hence, we will not consider these two cases in our analysis. For cases two and three, it is important to mention that it is essential that measurements cannot be made simultaneously on the 'same' sample. This means that either the sample where we measure changes with the measurement or it becomes inaccessible. Hence, we need to take a new sample for every next measurement. In fact, if all measurements can be made on the same sample, then the conditions of possible experience will always hold. Indeed, for macroscopic physics, we are able to create 'ensembles' which correspond to a large number of equivalent systems to which we can subject experimental measurements and thus calculate relative frequencies in a rather controlled way. The latter is exactly the reason why the conditions of possible experience hold in classical physics, and why we believe they are so important to explain reality. 

However, when we move to domains where multiple measurements cannot be made on the same sample, such as in quantum physics or in psychology, and where we see highly contextual situations that suggest intrinsic nondeterministic changes of state (physical or psychological, respectively), we shall carefully consider whether the conditions of possible experience are truly a constraint of what we are studying. Moreover, if they are not the right constrain, it is a scientific duty to explain what kind of restrictions could be in operation.

\section{How the conjunction fallacy data looks?}

We performed a literature review considering all articles citing SA between 1983 (year of publication) and 2016. Given the importance of the factual analysis of the previous section, we would have expected that studies on the CF generally report statistics of the values $P(A),P(B)$ and $P(AB)$ obtained from their experiments. However, we observed that studies reporting such data are rare. 
In table~\ref{data0} we show the number of articles citing SA, the number of works which produced experimental data on probability estimates, and the number of articles that additionally reported statistics of the values $P(A),P(B)$ and $P(AB)$.

\begin{table}[ht]
\centering
\begin{tabular}{|c|c|c|}
\hline
Number of articles & Experiments on probability estimates & Well reported data \\
\hline
4080 & 272 & 37 \\
\hline
\end{tabular}
\caption{Summary of CF literature and data.}
\label{data0}
\end{table}

A closer inspection to the reported data shows that the full extent of possible experiences, represented by $[0,1]^3$ has only been narrowly covered by current experiments. We compiled the complete set of data points (416 points) reported by the articles in the third column of table~\ref{data0} in order to see how well-distributed is the data across the set of possible points that can be obtained. We had to adapt some of these data points to the interval $[0,1]$ when they were reported on a different scale. In table~\ref{cover} we see the number of data points reported by experiments in different sub-volumes of $[0,1]^3$.

\begin{table}[ht]
\centering
\begin{tabular}{|c|c|c|c|}
\hline
$P(A)$ & $P(B)$ & $P(AB)$ & frequency \\
\hline
$[0,0.5]$ & $[0,0.5]$ & $[0,0.5]$ &115\\ \hline
$[0,0.5]$ & $[0,0.5]$ & $(0.5,1]$ & 2\\ \hline
$[0,0.5]$ & $(0.5,1]$ & $[0,0.5]$ & 37\\ \hline
$[0,0.5]$ & $(0.5,1]$ & $[0.5,1]$ & 10\\ \hline
$(0.5,1]$ & $[0,0.5]$ & $[0,0.5]$ & 106\\ \hline
$(0.5,1]$ & $[0,0.5]$ & $(0.5,1]$ & 41\\ \hline
$(0.5,1]$ & $(0.5,1]$ & $[0,0.5]$ & 3\\ \hline
$(0.5,1]$ & $(0.5,1]$ & $[0.5,1]$ & 102\\ \hline
\end{tabular}
\caption{Number of data points reported on each subvolume of $[0,1]^3$.}
\label{cover}
\end{table}

We observe that, at our volume resolution, all subvolumes have at least 2 data points. However, we see that the distribution is far from homogenous, implying that experiments have not properly sampled the space of possibilities. In particular, we observe that the second case is the most underrepresented case, which consists of $P(A)$ and $P(B)$ being smaller than $0.5$ while $P(AB)$ is larger than $0.5$. This situation is an example of what has been called a 'double overextension' in categorization~\cite{h1988a}, and refers to the conjunctive category being more typical (in the CF case, more likely) than the two former concepts (in the CF case, former events). The other most underrepresented case is the seventh case which reflects a possible experience in the Boole's sense. This also suggests that there has not been a systematic study in finding the frequency of CF in real situations. 
 
\begin{figure}[htbp]
  \centering
  \includegraphics[width=\textwidth]{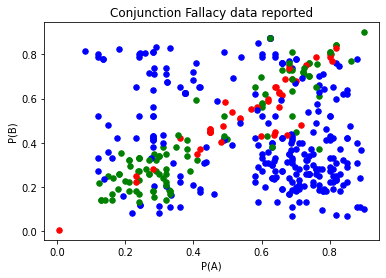}
  \caption{CF data reported and coloured by the extent fallacy violation. Green means no violation, blue means violation on one of the probabilities, and red means violation on two of the probabilities.}
  \label{points}
\end{figure}

To deepen a bit our analysis, we show in figure~\ref{points} the 416 data points, plotting $P(A)$ versus $P(B)$ and labelling the colour of the point based on the value of $P(AB)$ in relation to the other two. That is, for $P(AB)\leq \min(P(A),P(B))$ the point is colored green (no CF), for $\min(P(A),P(B))\leq P(AB)\leq\max(P(A),P(B)) $ the point is colored blue (single CF violation), and for $\max(P(A),P(B))\leq P(AB)$ the point is colored red (double CF violation).

Various things are worth noticing here. First, almost no data is reported for events with a probability smaller than $0.1$. This means that the phenomenology of very unlikely events is underinvestigated. Second, although points are relatively well distributed (with the exception of the very unlikely events mentioned before), there seems to be an area around $[0.5,0.6]\times[0,1]$ which is also under-represented. Interestingly, this area corresponds to events where there is very high uncertainty on $A$, and we observe from the few points available in that area that the three colours are more or less equally present. The latter is consistent with the fact that the more uncertainty about the events we have, the more diverse our thinking can be, which in turn is consistent with the fact that intuition operates more strongly in situations of uncertainty. Third, we see that both the green and red points are mostly scattered around the diagonal. This implies that green points are highly underrepresented, as we can always create macroscopic physical situations where $P(AB)$ is anywhere below both $P(A)$ and $P(B)$, and that 'red' deviations have probably not been explored enough.

\section{Conclusion: Do we need factual paradigms?}

In this article, we aim to illustrate that experimental research in CF has not given sufficient attention to what could be the most interesting question on the topic. Namely, what are our possible conjunctive experiences? 

In particular, we notice that research has focused on testing the existence of the fallacy under several different conditions, while identifying how often this happens, or to what extent it happens, has been left as secondary topics. By separating the data points reported in the literature into sub-volumes we see that the single fallacy, i.e. when $\min(P(A),P(B))< P(AB)\leq \max(P(A),P(B))$, is relatively well-explored, while the no fallacy ($P(AB)\leq \min(P(A),P(B))$) and the double fallacy ($\max(P(A),P(B)< P(AB))$) are under-explored (see table~\ref{cover}).

We believe that this lack of attention is linked to the rationalist approach to cognition. Namely, most authors have tried to explain 'why' people commit the fallacy by developing theories of how we interpret language and have focused on the commitment frequency of the fallacy in a given situation rather than on the extent at which occurs. However, conjunction fallacy explanations assume that if we would 'interpret the experiment text literally', we would not commit the fallacy. This is coherent with the fact that this phenomenon is called a fallacy, and it is not instead conceived as a 'equally valid' form of cognition as rationality.

From the sampled literature we see that potentially interesting cases with strong double fallacy such as $P(A)=0.1, P(B)=0.1$ and $P(AB)=0.5$ are either impossible or have not been explored. We call attention to this kind of situation because it exemplifies the extreme of our not rational cognition. For example, consider a typical romantic-story-like situation where two people meet by accident (in a train station for example) and must spend some little time together in the cold to catch their trains, and they fall in love in that short moment. We can clearly see that the chances that for two people 'their trains failing simultaneously' and 'falling in love in a small chat' can be rendered very small, but their conjunction is commonplace in our romantic culture, reflecting that we do believe that those events are quite possible, despite the rational fact that they are simply the conjunction of two unlikely events. 

Multiple ideas have been developed over the course of the previous century trying to face the domain of what we rationally believe as impossible. One of the most interesting of such attempts is the concept of synchronicity, worked out together between Pauli and Jung, prominent figures in quantum theory and psychology respectively~\cite{lindorff1995psyche}. They analyzed the idea that events, although not causally connected, can still occur simultaneously in a meaningful way. In other words, two or more seemingly unrelated events happen at the same time, and yet they have a non-causal underlying connection. When these two non-causally connected events are both unlikely is when synchronicity shows its most interesting power and it is where our culture has recognized in multiple stories, in many cases related to love, destiny, and other concepts where science has not been able to penetrate. Interestingly, synchronicity has been proposed to bridge the gap between quantum theory and psychology at a more fundamental level of reality where both kinds of events, psychological and physical, share the same nature~\cite{cambray2009synchronicity}. Along this same line, other researchers have attempted to prescind from the idea of theoretical constructions implying divisions between physics and psyche, and have attempted to develop a purely factual construction of reality, which cannot be disentangled from our perception. As such these attempts start directly from perceived reality and builds factual probabilistic laws based on our ability to replicate observational procedures~\cite{mugur2002objectivity}. 
To conclude, we propose that the conjunction fallacy must be analyzed under these perhaps more speculative lines to advance a deeper understanding of what it means for our cognition. Specifically, experiments exploring strong double fallacies could provide a much more solid ground to prove the rationalistic-inspired approaches incomplete. Firstly, if we had more properly designed studies, in line with our discussion of constructing CF test questions in section 4, we could say more clearly whether such double fallacies are truly unexplored or truly unlikely. As it stands, most of the studies we have analysed do not allow us to derive a logical and methodological conclusion on this point. Secondly, we would have a more rigorous argument to talk about our irrationality and that this irrationality is not fallacious, but normal for the behaviour of our brain, and thus factual approaches for constructing scientific knowledge such as~\cite{mugur2002objectivity} might prove specially useful, and might invite to explore less traditional mathematical structures. Concerning psychological experiments, with the proper design of the task, we would not lose the respondents' focus by dispersing them into ratings by considering the task in a too general way. On the contrary, an assignment in which respondents are asked to estimate the likelihood of each event (in the context of Linda's story or similar) allows them to analyse these events separately and make a more accurate estimate, which will give us prospectively more methodological power to justify the radically non-intuitive nature of CF.

\bibliographystyle{plain}
\bibliography{name}

\end{document}